\documentclass[10pt,twocolumn,letterpaper]{article}

\usepackage{iccv}
\usepackage{times}
\usepackage{epsfig}
\usepackage{graphicx}
\usepackage{amsmath}
\usepackage{amssymb}

\graphicspath{ {images/} }
\usepackage[utf8]{inputenc}
\usepackage[T1]{fontenc}

\usepackage{amsfonts}
\usepackage{booktabs}
\usepackage{siunitx}
\usepackage{makecell}
\usepackage{caption}
\usepackage{subcaption}

\usepackage[pagebackref=true,breaklinks=true,letterpaper=true,colorlinks,bookmarks=false]{hyperref}

\iccvfinalcopy 

\def\iccvPaperID{5600} 

\ificcvfinal\fi
\begin{document}

\title{Hybrid Cosine Based Convolutional Neural Networks}

\author{Adri\`a Ciurana, Albert Mosella-Montoro, Javier Ruiz-Hidalgo\\
Universitat Polit\`ecnica de Catalunya\\
Image Processing Group - Signal Theory and Communications\\
{\tt\small https://imatge.upc.edu} \\
{\tt\small adri00796@gmail.com, albert.mosella@upc.edu, j.ruiz@upc.edu}
}

\maketitle

\begin{abstract}
Convolutional neural networks (CNNs) have demonstrated their capability to solve different kind of problems in a very huge number of applications. 
However, CNNs are limited for their computational and storage requirements. 
These limitations make difficult to implement these kind of neural networks on embedded devices such as mobile phones, smart cameras or advanced driving assistance systems. 
In this paper, we present a novel layer named Hybrid Cosine Based Convolution that replaces standard convolutional layers using cosine basis to generate filter weights. 
The proposed layers provide several advantages: faster convergence in training, the receptive field can be increased at no cost and substantially reduce the number of parameters. We evaluate our proposed layers on three competitive classification tasks where our proposed layers can achieve similar (and in some cases better) performances than VGG and ResNet architectures.
\end{abstract}


\section{Introduction}
Convolutional neural networks (CNNs) are a category of neural networks that have achieved an extremely good performance in a multitude of tasks, for example in image classification~\cite{VGG,ResNet} and segmentation~\cite{FCNSemanticSegmentation, pspNet}. CNNs learn filters that allows the network to capture patterns in a feature space. During the last few years different alternatives to CNNs have been investigated such as \emph{MobileNets}~\cite{mobileNet} and \emph{Dilated Convolutions}~\cite{dilatedConvolution}. MobileNets use depth-wise and point-wise convolutions in order to compress neural networks and reduce the number of needed calculations. Dilated convolutions allows an expansion of the receptive field without loss of resolution or coverage.

On the other hand, Discrete Cosine Transforms (DCTs) have been widely used in many image processing applications such as the JPEG encoding standard in order to decompose an image into its spatial frequency spectrum. DCTs have also been applied into deep learning frameworks. For example, in~\cite{highSpeedDCT}, DCTs are used for speed up training of neural networks. Furthermore, \emph{Ghosh \etal}~\cite{deepFeatureDCT} demonstrate in their work that applying DCTs on the feature maps generated by layers of CNNs can result in a speed up in the converge of the training and, in some cases, an increment of the accuracy.

\begin{figure}[t]
    \centering
    \includegraphics[scale=0.35]{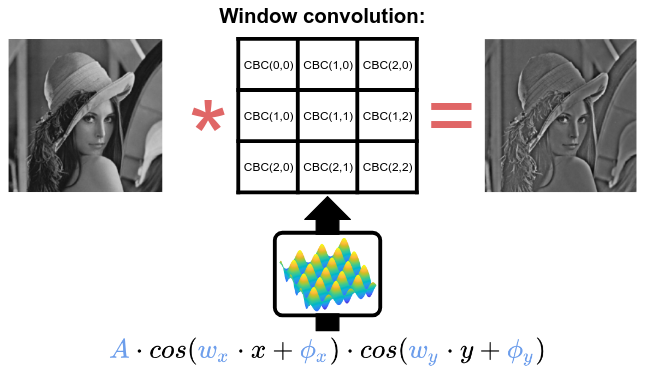}
    \caption{Pipeline to generate filter weights from the cosine basis of CBC filters before the convolution operator. $A$ denotes the amplitude, $w_x$ and $w_y$ the frequencies of the horizontal and vertical dimensions and $\phi_x$, $\phi_y$ the corresponding phases. A single channel is assumed in this example.}
    \label{fig:pipeline}
\end{figure}

In this paper, motivated by the advantages that DCTs (and frequency analysis in general) have provided on several areas such as image processing and deep learning, we propose a new variant of the standard convolution filter named Cosine Based Convolution filter or CBC. In this variant, instead of learning directly filter weights, frequencies of cosine basis that later will generate convolution weights are learned. Figure~\ref{fig:pipeline} shows a graphical overview of how CBC filters generate convolutional filter weights. Later on, we extend this filter to the entire layer using a hybrid approach where proposed CBC filters are used together with standard convolutional filters.

The main contributions of this paper are:
    
\begin{itemize}
\item A new variant of the standard convolutional layer, hybrid CBC layer, where filter weights are obtained from a cosine basis operation.
\item Hybrid CBC layers require less parameters as only amplitudes, frequencies and phases are needed to represent filter weights. 
\item CBC parameters are independent of the convolution filter size thus receptive fields can be increased without any increase at no cost. 
\item As less parameters need to be learned, hybrid CBC layers provide faster converge on training.
\end{itemize}

Experimental results show that even though less parameters are used, similar (and in some cases better) accuracy can be obtained from CBC layers in classification tasks with different datasets.

The paper is organized as follows. We review the related work in Section \ref{sec:rw}. In Section \ref{sec:cbc} the Cosine Based Convolutional filter is described. Following in Section \ref{sec:hcbc} the hybrid extension is explained. Section \ref{sec:ablation} describes the studies done for each variant of our proposed method and their performance. In Section \ref{sec:results}, the results of our proposal applied to different kind of datasets and architectures are showed. Finally conclusions are drawn in Section \ref{sec:conc}.


\section{Related work}\label{sec:rw}
During the last years, frequency analysis have been applied to improve several deep learning architectures. For example, \emph{Jadeberg \etal}~\cite{lowRankExpansion} showed a methodology that reduces memory usage and speeds up the inference stage. It consists of the approximation of the trained rank-N convolutional neural network filters by separable rank-1 filters. In an extension to this, the standard convolution on the spatial domain was replaced by multiplications of the transformed filters in the frequency domain~\cite{fastTrainFft}.

Hashed Networks exploit inherent redundancy in neural networks using a low-cost hash function to randomly group connection weights into hash buckets, in which all the weights share the same parameter value~\cite{hashingTrick}. An extension by \emph{Chen \etal}~\cite{compressingConvolutionalNeuralNetworks} groups the weights based on their DCT representation. Another methodology by \emph{Han \etal}~\cite{deepCompression} prunes the network by learning only the important connections. Then the weights are quantified to enforce weight sharing. Finally, Huffman encoding is applied.

In a more close relation to the work proposed in this paper, \emph{Wang \etal}~\cite{cnnPack} present a method to compress neural networks in the frequency domain. Using a DCT, filter weights are clustered. These weights are quantized and encoded using Huffman coding. However, this compression is only applied for storage purposes.

In a similar manner, \emph{Uliný \etal}~\cite{harmonicNetworks} present harmonics blocks that are used to replace standard convolutional layers. Harmonic blocks consist of a convolution with a filter bank that isolates the coefficients of the DCT basis functions to their exclusive feature maps, creating a new feature map per each channel and each frequency defined by hand. However, the spectral decomposition of this proposal upsamples the number of intermediate features between layers thus notably increasing the corresponding memory requirements. In our case, DCT frequencies are learned so only the more relevant decomposition are used in the network.

\section{Cosine Based Convolution filter (CBC)}\label{sec:cbc}
This section describes the proposed CBC filter. This filter modifies the standard convolution by using cosine basis to generate filter weights (that will be applied in a convolution operator). Therefore, CBC filters do not learn directly filter weights but the frequency parameters (amplitudes, frequencies and phases) that will allow to generate the filter weights in the spatial domain (the domain where the convolution operation works). In this work, the cosine basis of CBC filters can be interpreted as a frequency decomposition of convolutional filter weights (not input images or features). The main advantages of the CBC filters are that frequency parameters represent convolutional filter weights in a more compressed representation and that they are invariant to the window size of the convolution filter. 

Figure~\ref{fig:pipeline} represents the pipeline to generate convolution filter weights from the cosine basis defined by the CBC layer. As it can be seen, the same cosine basis can represent convolutional filter weights for different filter sizes without the need of extra parameters.

In order to define CBC filters, we define different options to generate convolutional filter weights in the spatial dimensions, $(x,y)$, and in the channel/feature dimension, $c$, as frequency harmonics are more common in spatial dimensions in natural images. Therefore, we will separate the cosine basis to represent convolutional filter weights in the spatial dimension and in the feature dimension. Note that these spatial and feature dimensions refer to the convolutional filter weights and not the image or signal it is convolved with. This section assumes 2D images and 2D convolutional operations but higher dimensions can easily be extrapolated. 

\subsection{Spatial dimensions}

\begin{figure}
    \centering
    \begin{subfigure}[b]{0.49\columnwidth}
        \centering  
        \includegraphics[width=\linewidth]{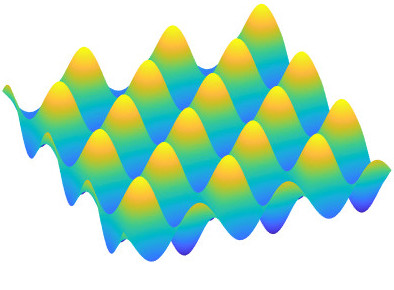} 
        \caption{spatial-product ($S_{P}$)}
        \label{fig:lc}
    \end{subfigure}
    \hfill
    \begin{subfigure}[b]{0.49\columnwidth}
    
        \centering
        \includegraphics[width=\linewidth]{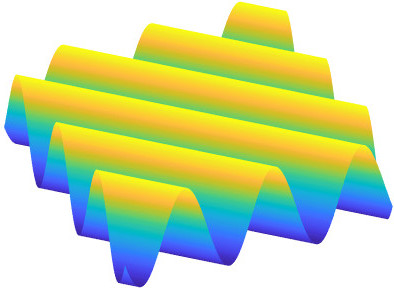}
        \caption{spatial-direction ($S_{D}$)}
        \label{fig:uc}
    \end{subfigure}
\caption{Comparison the degrees of freedom between the spatial-product and the spatial-direction.}
\label{fig:cbc_spatial}
\end{figure}

We propose two different cosine basis to generate filter weights on the spatial dimension of the filter. The first basis, which we called \emph{spatial-product} or $S_{P}(x,y)$, is built with the product of two cosines where each of them represents a unique frequency, $w_x$ and $w_y$, in the vertical and horizontal dimensions respectively:

\begin{equation}
    S_{P}(x,y) = cos(w_x \cdot x + \phi_x) \cdot cos(w_y \cdot y + \phi_y)
    \label{eq:spatial-product}
\end{equation}
where $x$ and $y$ represent the spatial coordinate of convolutional filter weights. 

Even though this cosine basis will allow the generation of a large spectrum of different signals (filters), it only allows the composition of two unique spatial harmonics and directional filters are not represented. For this reason we propose and alternative basis, which we called \emph{spatial-direction} or $S_{D}(x,y)$,  where only one harmonic is represented but different 2-dimensional directions can be achieved:

\begin{equation}
    S_{D}(x,y) = cos(w_x \cdot x + w_y \cdot y + \phi)
    \label{eq:spatial-direction}
\end{equation}

Figure~\ref{fig:cbc_spatial} represents the different degrees of freedom between the two cosine basis proposed for the spatial dimensions. It can be seen that \emph{spatial-direction} is able to generate filters on different directions using a single harmonic while \emph{spatial-product} represents two harmonics in the vertical and horizontal dimensions.

\subsection{Feature dimension}
In case of the feature dimension, we also propose two different alternatives.The first basis, called \emph{feature-direct} or $F_{D}(c)$, is built as a cosine with a new frequency $w_c$ and phase $\phi_c$ along the feature dimension of the filter:

\begin{equation}
    F_{D}(c) = A \cdot cos(w_c \cdot c + \phi_c)
    \label{eq:feature-direct}
\end{equation}
where $A$ denotes the amplitude of the cosine basis and $c$ represents the feature coordinate of convolutional filter weights.

In order to relax the previous restriction, we introduce a second alternative, called \emph{feature-weight} or $F_{W}(c)$, without any cosine basis but directly using a different amplitude $A(c)$ for each input channel (as opposed to $A$ in Eq.~\eqref{eq:feature-direct} that uses the same amplitude for all channels): 

\begin{equation}
    F_{W}(c) = A(c)
    \label{eq:feature-weight}
\end{equation}

Using \emph{feature-direct} offers higher compression in the number of parameters of the CBC layer as only 3 parameters (amplitude $A$, frequency $w_c$ and phase $\phi_c$) can represent $C$ channels in the feature dimension (in fact, they are completely independent of the convolutional filter size). In contrast, it forces an order on the feature dimension and the optimizer should not only find the best parameters, but should find the order of the features to obtain the best performance. Using \emph{feature-weight}, a different weight $A(c)$ for each feature coordinate (channel) is used and therefore the number of parameters in not reduced in the feature dimension but the complexity of the optimizer decreases.

\subsection{CBC filter}
Finally, both spatial and feature basis can be combined to create the CBC filter that represent convolutional filter weights. We combine spatial dimensions using the same frequency for all input channels:

\begin{equation}
    CBC(x,y,c) = \underbrace{S(x,y)}_{\text{Spatial dimensions}}  \cdot \underbrace{F(c)}_{\text{Feature dimension}}
    \label{eq:single-channel}
\end{equation}

where $S(x,y)$ represents any of the two spatial domain cosine basis, \emph{spatial-product} ($S_P(x,y)$ in Eq.~\eqref{eq:spatial-product}) or \emph{spatial-direction} ($S_D(x,y)$ in Eq.~\eqref{eq:spatial-direction}) and $F(c)$ represents any of the two feature domain basis, \emph{feature-direct} ($F_D(c)$ in Eq.~\eqref{eq:feature-direct}) or \emph{feature-weight} ($F_W(c)$ in Eq.~\eqref{eq:feature-weight}). For example, using \emph{spatial-product} for the spatial dimension and \emph{feature-direct} for the feature dimension the following cosine basis will be used to represent the filter convolutional weights: 

\begin{equation}
    A \cdot cos(w_x \cdot x + \phi_x) \cdot cos(w_y \cdot y + \phi_y) \cdot cos(w_c \cdot c + \phi_c)
    \label{eq:cbc_example}
\end{equation}

In this case, only 7 parameters are needed to represent the cosine basis: the amplitude $A$, frequencies $(w_x, w_y, w_c)$ and phases $(\phi_x, \phi_y, \phi_x)$ while it can represent up to $H{\times}W{\times}C$ weights of a standard convolution filter (where $H$ if the kernel height, $W$ the kernel width and $C$ the number of input channels). 

Combining all cosine basis proposed, we obtain up to 4 variations for the CBC filter, depending on the different options for the spatial and feature dimensions. Table~\ref{table:cbc_combinations} summarizes all possible CBC filter combinations (and their respective equations) together with the number of parameters needed to represent the convolution filters.

Using the same spatial frequency for all channels provides the most efficient parameter compression. Using different spatial frequencies for each channel could provide a higher capability of representation for CBC layers. However, our initial search showed that using the same spatial frequencies is enough to capture a similar feature representation of standard convolutional layers and are the only ones considered in this paper. 

\begin{table*}[!ht]
\footnotesize	
\begin{center}
\scalebox{0.9}{
\begin{tabular}{| c | c | c | c | l | l |}
\hline
Name & Spatial dimensions & Feature dimension & Parameters & Equation \\
\hline \hline
2D conv   & -                        &                     - & $H{\times}W{\times}C$ & $W_{x,y,c}$ \\
CBC-$S_PF_D$ & \emph{spatial-product}   & \emph{feature-direct} & $1 + 2 + 2 + 2$           & $S_{P}(x,y) \cdot F_{D}(c) =A \cdot cos(w_x \cdot x + \phi_x) \cdot cos(w_y \cdot y + \phi_y) \cdot cos(w_c \cdot c + \phi_c)$ \\
CBC-$S_PF_W$ & \emph{spatial-product}   & \emph{feature-weight} & $C + 2 + 2$           & $S_{P}(x,y) \cdot F_{W}(c) = A(c) \cdot cos(w_x \cdot x + \phi_x) \cdot cos(w_y \cdot y + \phi_y)$ \\
CBC-$S_DF_D$ & \emph{spatial-direction} & \emph{feature-direct} & $1 + 3 + 2$               & $S_{D}(x,y) \cdot F_{D}(c) = A \cdot cos(w_x \cdot x + w_y \cdot y + \phi) \cdot cos(w_c \cdot c + \phi_c)$\\
CBC-$S_DF_W$ & \emph{spatial-direction} & \emph{feature-weight} & $C + 3$  & $S_{D}(x,y) \cdot F_{W}(c) = A(c) \cdot cos(w_x \cdot x + w_y \cdot y + \phi)$ \\
\hline
\end{tabular}
}
\end{center}
\caption{CBC filter combinations and the number of parameters needed to represent $H{\times}W{\times}C$ convolutional filter weights.}
\label{table:cbc_combinations}
\end{table*}


\section{Hybrid CBC Layer}\label{sec:hcbc}
\begin{figure}[ht]
    \centering
    \includegraphics[scale=0.3]{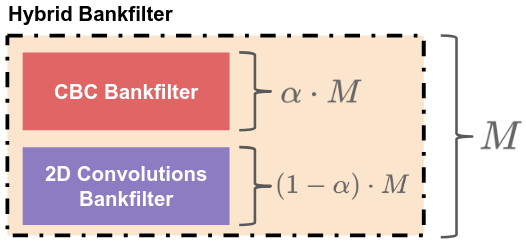}
    \caption{Representation of a hybrid CBC layer with $M$ total filters. $\alpha$ controls the number of CBC filters with respect conventional convolutions.}
    \label{fig:hybrid_layer}
\end{figure}

Even though all convolution filters of a CNN layer could be replaced with CBC filters, we propose an hybrid CBC layer that combines the best of both worlds: conventional convolution filters and CBC filters. In this way, the hybrid layer can use CBC filters (with less parameters) to efficiently represent harmonics in the input signal while it can also use conventional convolution filters (with more parameters and more degrees of freedom) to represent more complex features of the input signal. Figure~\ref{fig:hybrid_layer} shows a representation of a HCBC layer with $M$ total filters. The parameter $\alpha$ controls the number of CBC filters with respect to the conventional convolutions. If $\alpha$ is close to $1$ then the HCBC layer will be composed mostly of CBC filters while $\alpha$ close to $0$ will create hybrid CBC layers that behave as standard convolutional layers. Section~\ref{sec:ablation} will study the effect of using different $\alpha$ values in HCBC layers.

Note that in the specific case of $1{\times}1$ convolution layers, using a cosine basis in the spatial domain does not provide any gain as only $1$ parameter is used per channel. Therefore, we set $S(x,y)=1$ in Eq.~\eqref{eq:single-channel} and only the feature dimension $F(c)$ is used. 


\section{Ablation studies}
\label{sec:ablation}
In this section we will analyze the different options proposed in this paper that are summarized in Table~\ref{table:cbc_combinations}. We will use two known neural network architectures, VGG~\cite{VGG} and ResNet~\cite{ResNet}, where all their convolutional layers are replaced by hybrid CBC layers with different options.  

We have selected VGG16bn and ResNet50 architectures as the specific architectures for VGG and ResNet for our experiments because: a) they are widely used by industry and academia, b) they provide a very robust architecture with very good results in many applications and c) they have a very different internal architecture.   

For the ablation studies, we use the CIFAR10~\cite{krizhevsky2009learning} dataset and show results for both the proposed hybrid CBC layers and the network architectures without modification which we consider baseline. All studies will show the loss curves for the training and validation sets for $200$ epochs (graphs in the first row of Figures). In order to facilitate the comparison we also show the same loss with respect to the baseline (graphs in the second row of the Figures). Finally, we also incorporate a histogram that indicates the number of parameters used by each architecture in log scale (bottom row of Figures).

\subsection{Spatial criteria}

\begin{figure}[t]
    \centering
    \includegraphics[width=1\linewidth]{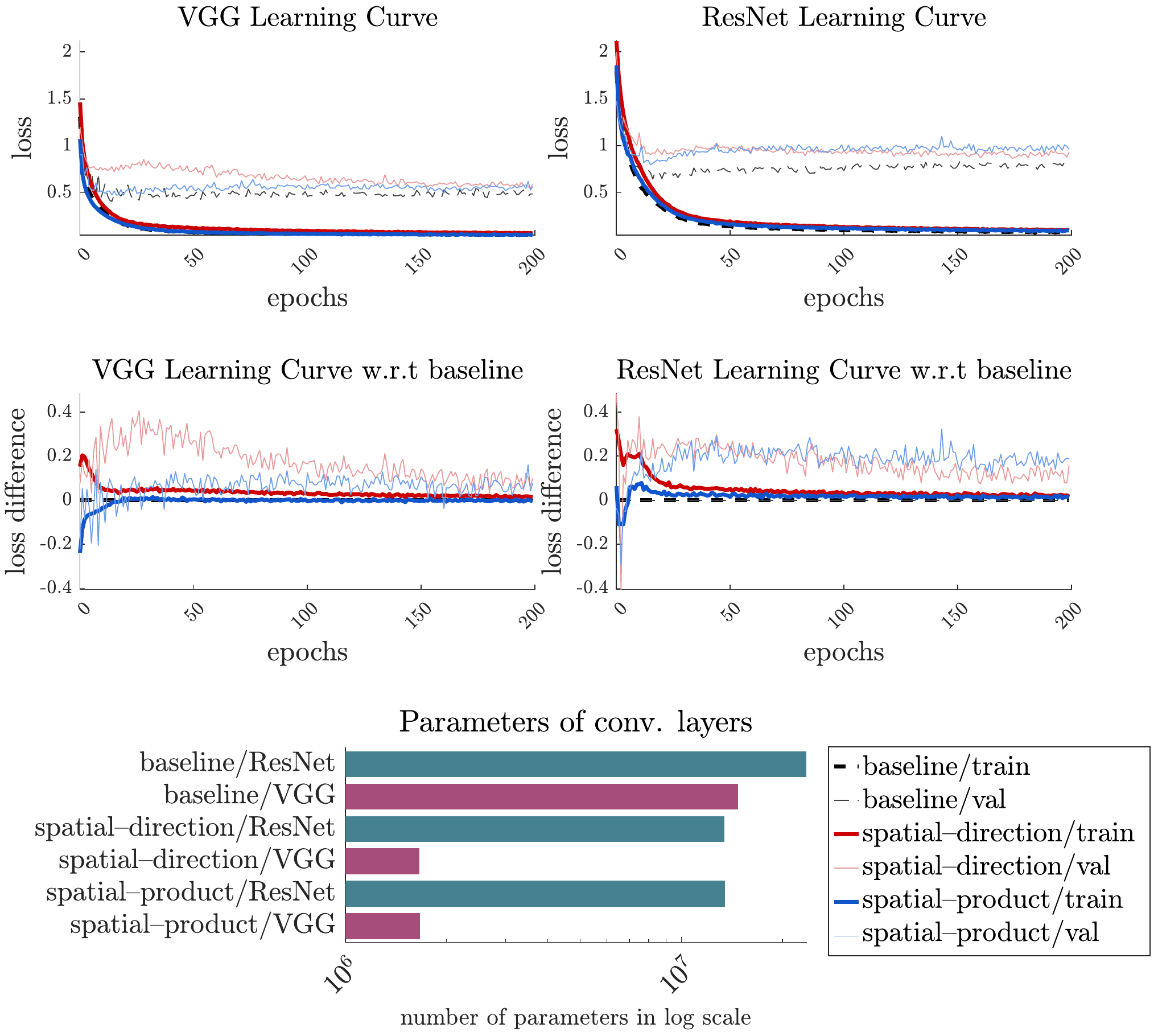}
    \caption{Results for spatial dimension studies using VGG and ResNet architectures. Best seen in color.}
    \label{fig:spatial_criteria}
\end{figure}

In this study, we analyze the behaviour of spatial dimension options, \emph{spatial-product} and \emph{spatial-direction}, during training. In order to focus only in the spatial dimension options, we fix the feature dimension to \emph{feature-weights}. Figure~\ref{fig:spatial_criteria} shows the learning curve of the different spatial options.

As it can be seen, in case of VGG, both spatial alternatives offer a similar behaviour. Both achieve similar performance than baseline but with a small penalty in convergence speed in the case of \emph{spatial-direction}. However, both achieve a great reduction of network parameters with a compression factor around $8.85$ with respect to the total parameters of convolutional layers in VGG. As will be stated in the experimental section, we measure the compression factor as the ratio between the total number of parameters in convolutional layers of baseline architectures (VGG or ResNet) with respect to the total number of parameters in CBC layers.

In case of ResNet architecture both spatial alternatives offer a similar performance. Neither \emph{spatial-direction} or \emph{spatial-product} are able to achieve the same performance than the baseline. Nevertheless it is very close to baseline with the advantage of the reduction of the number of parameters of the network with a compression factor around $1.75$ with respect to ResNet.

\subsection{Feature criteria}

\begin{figure}[!tb]
    \centering
    \includegraphics[width=1\linewidth]{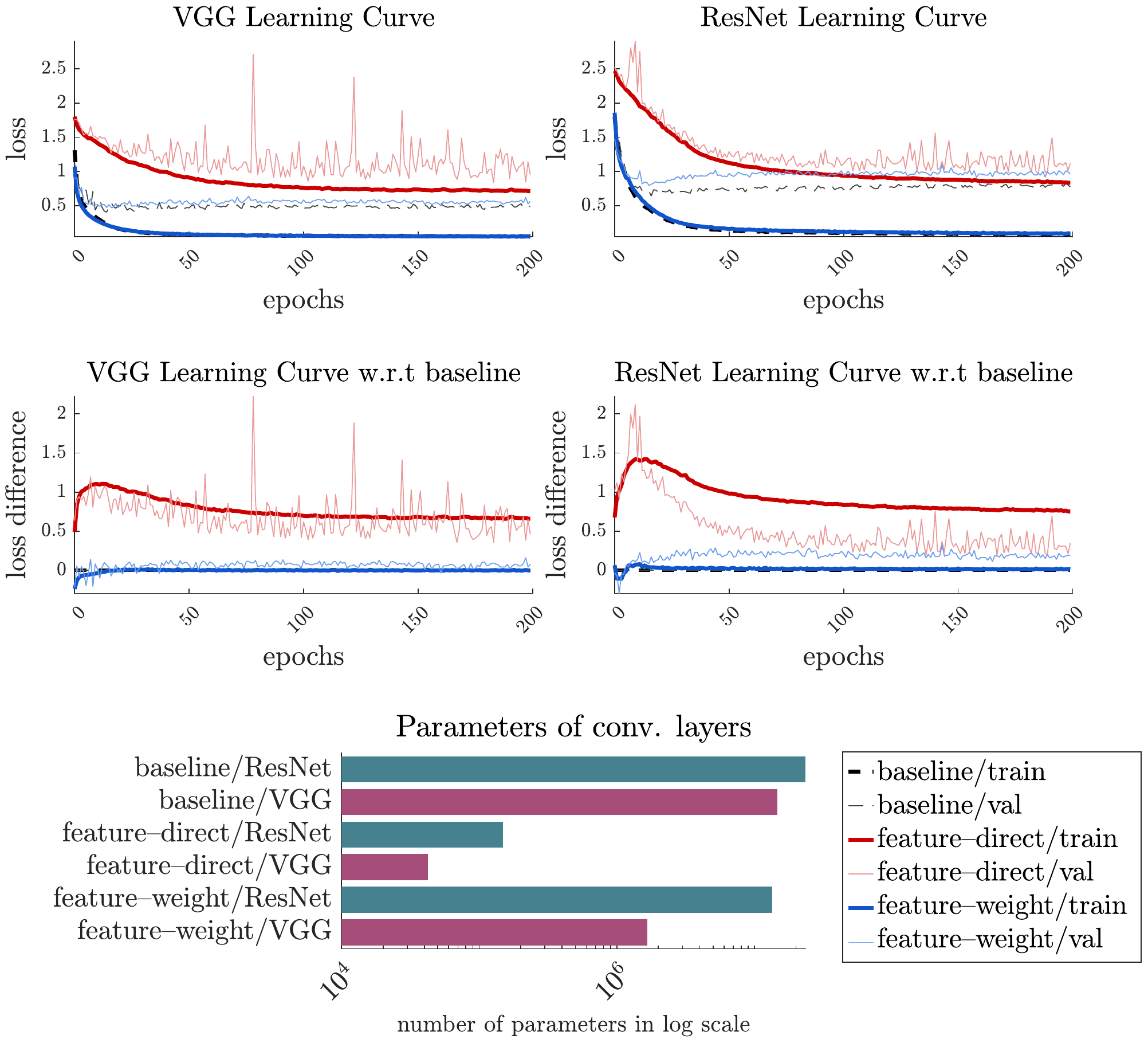}
    \caption{Results for feature dimension studies using VGG and ResNet architectures. Best seen in color.}
    \label{fig:feature_criteria}
\end{figure}

In this study, we analyze the behavior of feature dimension options, \emph{feature-direct} and \emph{feature-weight}, during training. For this, we will fix the spatial dimension as \emph{spatial-product}. Figure~\ref{fig:feature_criteria} shows the learning curve comparing both feature options.

It can be seen that for both architectures option \emph{feature-weight} has a much better behavior than \emph{feature-direct}. Even though \emph{feature-direct} has a higher compression ratio ($348.55$ for VGG and $158.66$ for ResNet), it cannot achieve a performance similar to baseline. This is due to the fact that combining channels using a cosine basis does not provide the flexibility needed by the network to learn the classification task. However, \emph{feature-weight} shows a behavior close to baseline with a compression ratio of $8.85$ for VGG and $1.75$ for ResNet.

\subsection{Hybrid CBC layer}

In this study, we analyze the performance of the hybrid CBC layer for different $\alpha$ parameters. We will test when $\alpha = 1$ (only CBCs filters) and $\alpha=0.5$ (half the filters of the layer will be standard convolutional filters and the other half CBC filters). For this analysis we have fixed the spatial dimension to \emph{spatial-product} and compare both \emph{feature-direct} and \emph{feature-weight} options. 

\begin{figure*}[ht]
    \centering
    \begin{subfigure}[b]{1\columnwidth}
        \centering
        \includegraphics[width=1\linewidth]{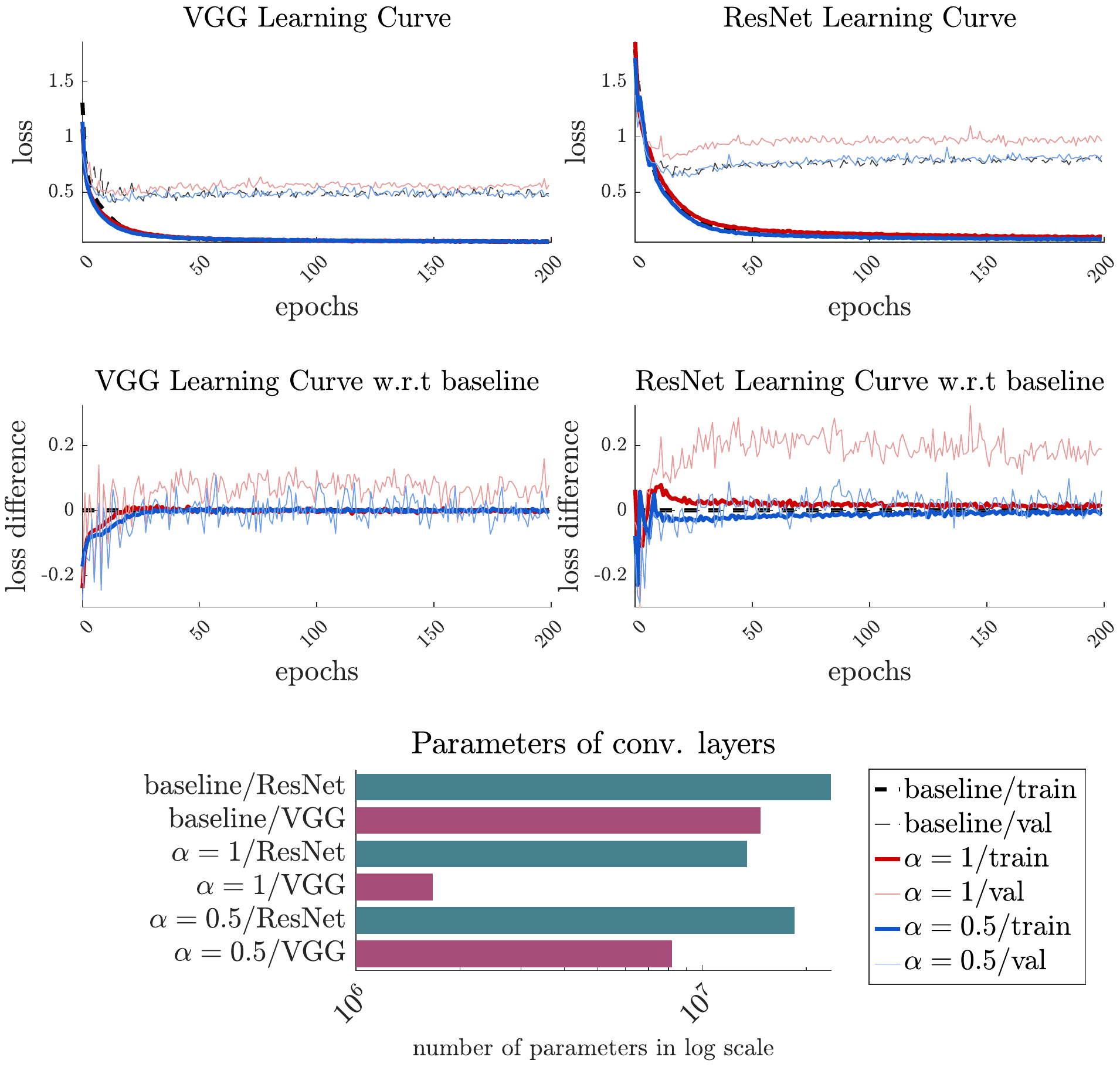}
        \caption{Feature-weight}
        \label{fig:hybrid_criteriav2}
    \end{subfigure}
    \hfill
    \begin{subfigure}[b]{1\columnwidth}
        \includegraphics[width=1\linewidth]{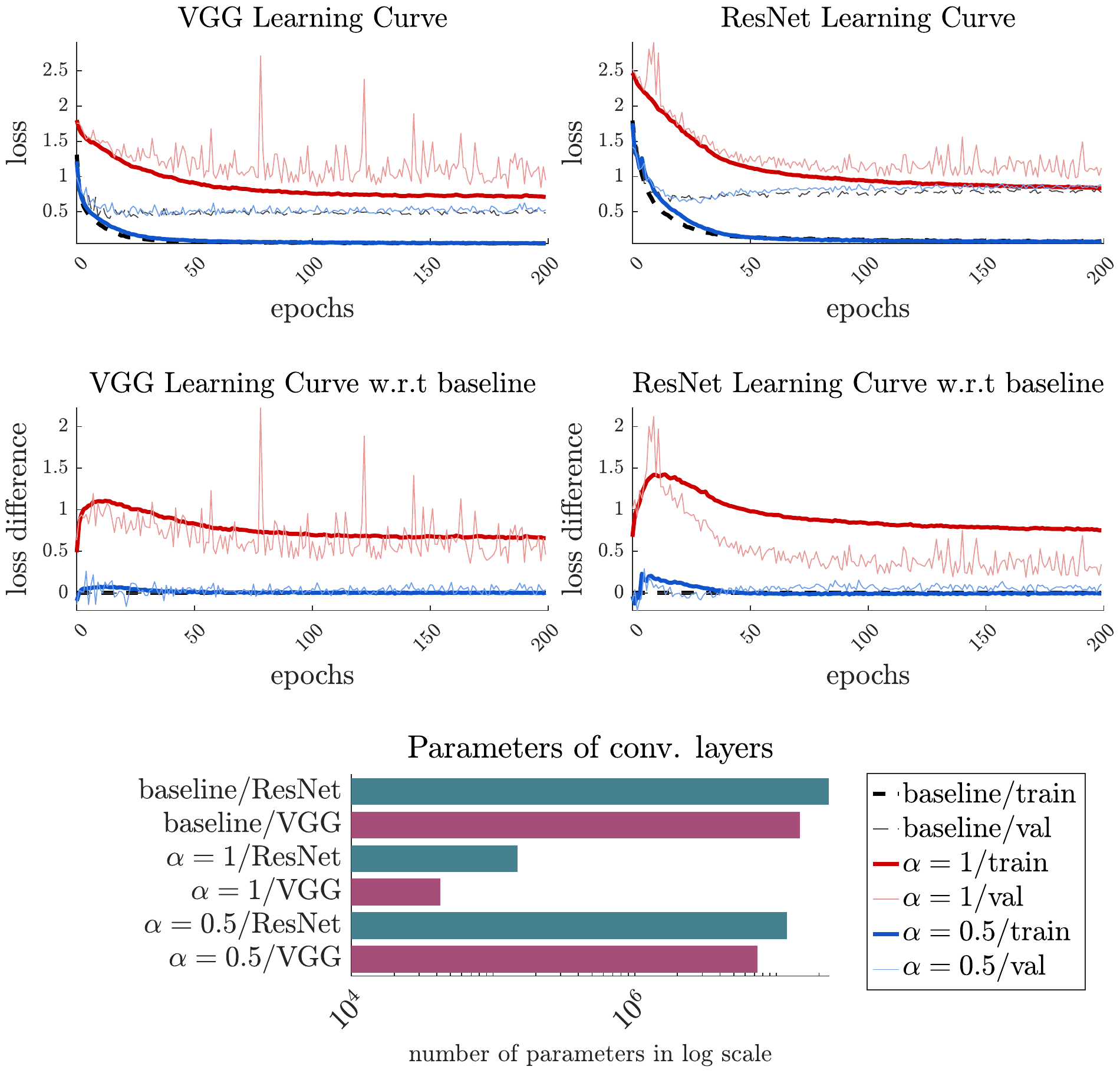}
        \caption{Feature-direct}
        \label{fig:hybrid_criteriav1}
    \end{subfigure}
\caption{Results for Hybrid-CBC layers on VGG and ResNet architectures with different $\alpha$ values. Best seen in color.}
\label{fig:cbc_v1_v2}
\end{figure*}

Figure~\ref{fig:hybrid_criteriav2} shows the behaviour for both $\alpha$ values when using the \emph{feature-weight} option for the feature dimension. Both VGG and ResNet architectures obtain very similar results for the two values of $\alpha$. When $\alpha=0.5$ the generalization of the networks is better than when $\alpha=1$. Moreover, when $\alpha=0.5$, the results are very close to baseline. This experiment results in compression factors of $1.79$ and $1.27$ for VGG and ResNet respectively.

Figure~\ref{fig:hybrid_criteriav1} shows the behaviour for the two values of $\alpha$ setting \emph{feature-direct} option for the feature dimension. In this case we can observe a huge difference when the value of $\alpha$ changes. The network is not able to reach the performance of the baseline when $\alpha=1$ for both architectures. However, when $\alpha = 0.5$, the network is able to have a similar performance than the baseline with compression factors of $1.99$ and $1.98$ for VGG and ResNet respectively. In this case, letting half the filters of the layer to remain as standard convolution filters helps to compensate the lower capacity of representation of CBC filters with the \emph{feature-direct} option.

To summarize, when $\alpha=1$, \emph{feature-direct} is not able to reach the same performance as baseline. However, if $\alpha = 0.5$, both \emph{feature-direct} and \emph{feature-weight} configurations achieve similar performance to baseline with the advantage of the compression ratios mentioned above.  

\section{Results}
\label{sec:results}
This section compares the most representative options of hybrid CBC layers against VGG and ResNet architectures for different challenging datasets. We have tried to use datasets with different resolutions and number of classes to assess the performance of the proposed network. We have used validation accuracy in a classification tasks to measure the performance of the different options. 

\subsection{Datasets}
The datasets used in these experiments are:

\begin{itemize}
    \item \emph{CIFAR10}~\cite{cifar10} used previously in ablation studies. This dataset contains a total of $60000$ color images of resolution $32{\times}32$. It contains $10$ classes uniformly distributed representing different animals and vehicles.
    \item \emph{CIFAR100}~\cite{cifar100} contains $60000$ color images of resolution $32{\times}32$. It contains $100$ classes uniformly distributed. This dataset present a more challenging classification problem than \emph{CIFAR10} as it includes a greater number of classes and a smaller number of samples per class ($600$ per class). We have select this dataset for its complexity in image classification task.
    \item  \emph{Monkeys}~\cite{monkeys} contains approximately $1400$ color images of resolution $400{\times}300$. It contains a total of $10$ classes representing a fine-grain classification of different monkey species. The dataset is uniform distributed and corresponds to approximately $140$ images per class. We have selected this dataset to assess the performance of the hybrid CBC layers on higher resolution images and with a lower number of examples per class.
\end{itemize}

\subsection{Dataset preparation}
In the case of the {CIFAR10/100} datasets, we have preserved the original image size of $32{\times}32$ (in the following section~\ref{sec:model_surgery} the modifications of the network to accept this resolution will be explained). In the case of Monkeys, input images of resolution $224{\times}224$ are used. We also applied online data augmentation. In the case of {CIFAR10/100} random horizontal flips with probability $p=0.5$ are used. In the case of Monkeys, as there are few images per class, we decided to apply: a) random rotations between $0$ and $45$ degrees, b) random crops of $224{\times}224$ and c) horizontal flips with probability $p=0.5$.

\subsection{Model Surgery}
\label{sec:model_surgery}
We have adapted the VGG architecture to work with images of resolution $32{\times}32$ as input for the CIFAR10/100 datasets. In the case of ResNet, changes are not necessary since a large part of its architecture is composed of convolutional filters and it has an average global pooling at the end that unifies the size. However, for VGG, we need to replace the last fully connected layers with a new fully connected layer of $512 \cdot num\_classes$ weights as it is usually done in the literature. This change modifies the total number of parameters of the VGG network. However, we tried to analyze the behaviour of replacing standard convolutional layers for hybrid CBC layers. Therefore, we will focus only on the number of parameters on convolutional layers and not fully connected layers. 

\subsection{Training parameters}
We have set a batch size of $128$ for {CIFAR10/100} and $32$ for Monkeys. We have used the Adam optimizer with an initial learning rate of $0.001$, $\beta_1 = 0.9$ and $\beta_2 = 0.999$. The selection of this optimizer is to simplify the gradient descent and let the optimizer choose the most suitable learning rate during training. The learning window has been set for $200$ epochs in {CIFAR10/100} and $300$ in Monkeys to limit the time of the experiments. Please note that for this experiments we are not achieving state-of-the-art results for the classification tasks as we want to compare the behaviour of the proposed hybrid CBC layers with respect to the original architectures. All networks are trained from scratch from initial random parameters.

\subsection{Implementation}
The implementation of the proposed layers has been carried out using the PyTorch framework~\cite{pytorch}. All the experiments have been carried out in GPU together with CUDA and cudNN to accelerate the learning process.

\subsection{Experiments}
We compare against untouched VGG and ResNet architectures (that we consider baselines). For our proposed options, we have replaced all convolutional layers in VGG and ResNet for hybrid CBC layers (note that for $1{\times}1$ convolution layers only the feature dimension is replaced in CBC layers). For simplicity and as the ablation studies suggested, we fixed \emph{spatial-product} for the cosine basis on the spatial dimension and we compare both \emph{feature-direct} and \emph{feature-weight} for the feature dimension. In both cases the hybrid CBC layers are used with $\alpha = 0.5$. In order to better evaluate the proposed method we will indicate the following for all cases:

\begin{itemize}
    \item \emph{Best Acc.}: Indicates the best accuracy obtained in the whole learning window ($200$ epochs for CIFAR10/100 and $300$ epochs for Monkeys) for the validation set.
    \item \emph{Epoch exceeds x\% Acc.}: Indicates in which epoch the validation accuracy has been exceeded. This allows us to see the learning speeds of the different architectures.
    \item \emph{Parameters of conv. layers}: Total number of parameters in the convolutional layers. 
    \item \emph{Compression factor}:  Ratio of compression between the total number of parameters in the convolutional layers of baseline with respect to the total number of parameters in hybrid CBC layers.
\end{itemize}

\subsubsection{CIFAR10}
Table~\ref{table:cifar10_table} shows the results obtained for the CIFAR10 dataset. It can be seen that in both VGG and ResNet architectures, hybrid CBC layers with \emph{feature-weight} are able to slightly exceed the accuracy of the baseline using nearly half of the convolutional parameters. Additionally, they are able to reach the proposed accuracy limit in fewer epochs as the amount of parameters to learn is reduced.

\begin{table}[ht]
\renewcommand{\arraystretch}{1.2}
\tiny
\centering
\noindent
\begin{tabular}{| l | c | c | c | c |}
\hline
Architecture & Best Acc. & \makecell{Epoch exceeds\\ 75\% Acc.} &  \makecell{Parameters\\of conv. layers} & \makecell{Compression\\factor} \\
\hline
\hline
VGG           & $88.48$          & $3$          & $14723136$         & $1.00$         \\
\hline
VGG-CBC-$S_PF_D$  & $88.15$          & $5$          & $\mathbf{7382688}$ & $\mathbf{1.99}$ \\
\hline
VGG-CBC-$S_PF_W$ & $\mathbf{88.62}$ & $\mathbf{2}$ & $8195712$          & $1.79$          \\
\hline
\hline
ResNet           & $81.59$          & $12$          & $23508032$          & $1.00$         \\
\hline
ResNet-CBC-$S_PF_D$  & $80.90$          & $16$          & $\mathbf{11828096}$ & $\mathbf{1.98}$ \\
\hline
ResNet-CBC-$S_PF_W$ & $\mathbf{81.75}$ & $\mathbf{10}$ & $18483136$          & $1.27$          \\
\hline
\end{tabular}
\caption{Results for the validation set of CIFAR10 dataset. In bold the best results for each column option.}
\label{table:cifar10_table}
\end{table}

\subsubsection{CIFAR100}

Table~\ref{table:cifar100_table} shows the results obtained for the CIFAR100 dataset. This dataset presents a higher challenge and validation accuracy is much lower than with other datasets. In this case, only in the ResNet architecture, hybrid CBC layers with \emph{feature-weight} are able to slightly outperform the baseline. Hybrid CBC layers on VGG architecture are slightly below baseline on validation accuracy. As commented previously, all of the options are able to reach the proposed accuracy in fewer epochs. Note that as the network architecture do not change between datasets, compression factors remain the same through the different experiments. 


\begin{table}[ht]
\renewcommand{\arraystretch}{1.2}
\tiny
\centering
\noindent
\begin{tabular}{| l | c | c | c | c |}
\hline
Architecture & Best Acc. & \makecell{Epoch exceeds\\ 45\% Acc.} &  \makecell{Parameters\\of conv. layers} & \makecell{Compression\\factor} \\
\hline
\hline
VGG           & $\mathbf{60.95}$          & $7$          & $14723136$         & $1.00$         \\
\hline
VGG-CBC-$S_PF_D$  & $58.95$          & $9$          & $\mathbf{7382688}$ & $\mathbf{1.99}$ \\
\hline
VGG-CBC-$S_PF_W$ & $60.78$ & $\mathbf{4}$ & $8195712$          & $1.79$          \\
\hline
\hline
ResNet           & $49.35$          & $17$          & $23508032$          & $1.00$         \\
\hline
ResNet-CBC-$S_PF_D$  & $48.78$          & $18$          & $\mathbf{11828096}$ & $\mathbf{1.98}$ \\
\hline
ResNet-CBC-$S_PF_W$ & $\mathbf{50.79}$ & $\mathbf{13}$ & $18483136$          & $1.27$          \\
\hline
\end{tabular}
\caption{Results for the validation set of CIFAR100 dataset. In bold the best results for each column option.}
\label{table:cifar100_table}
\end{table}

\subsubsection{Monkeys}

Table~\ref{table:monkeys_table} shows the results obtained for the Monkeys dataset. This dataset uses higher resolution images as we wanted to assess the adaptability of our proposed cosine basis to higher spatial dimensions in the input images. It can be seen that in both VGG and ResNet architectures, hybrid CBC layers with \emph{feature-weight} are able to improve the accuracy of baselines even though compression ratios of $1.79$ and $1.27$ are used. Similar to previous experiments, training convergence also improves with respect to baseline.

\begin{table}[ht]
\renewcommand{\arraystretch}{1.2}
\tiny
\centering
\noindent
\begin{tabular}{| l | c | c | c | c |}
\hline
Architecture & Best Acc. & \makecell{Epoch exceeds\\ 75\% Acc.} &  \makecell{Parameters\\of conv. layers} & \makecell{Compression\\factor} \\
\hline
\hline
VGG           & $72.40$          & $-$          & $14723136$ & $1.00$         \\
\hline
VGG-CBC-$S_PF_D$  & $81.34$          & $234$          & $\mathbf{7382688}$ & $\mathbf{1.99}$ \\
\hline
VGG-CBC-$S_PF_W$ & $\mathbf{86.87}$ & $\mathbf{96}$ & $8195712$          & $1.79$          \\
\hline
\hline
ResNet           & $89.74$          & $49$          & $23508032$          & $1.00$         \\
\hline
ResNet-CBC-$S_PF_D$  & $89.55$          & $41$          & $\mathbf{11828096}$ & $\mathbf{1.98}$ \\
\hline
ResNet-CBC-$S_PF_W$ & $\mathbf{93.36}$ & $\mathbf{24}$ & $18483136$          & $1.27$          \\
\hline
\end{tabular}
\caption{Results for the validation set of Monkeys dataset. In bold the best results for each column option.}
\label{table:monkeys_table}
\end{table}

\section{Conclusions}
\label{sec:conc}
In this work we have proposed a new set of Convolutional Neural Network compression techniques, called Hybrid Cosine Based Convolutional Neural Networks, which use a cosine basis to represent the weights of the convolutional  filters. In our experiments in classification accuracy, architectures based on hybrid CBC layers are able to obtain better performances that VGG and ResNet architectures even though they use less parameters in the convolutional layers. Compression factors between $1.2$ and $2$ can be used with similar or even better performances. Hybrid CBC layers provide faster convergence in training and receptive fields can be increased at no cost as the number of parameters does not change. Furthermore, the use of hybrid CBC layers is a straightforward replacement of current convolutional layers and no further adaptation is needed. In future work, we would like to study the applicability of hybrid CBC layers as feature detectors for other tasks, such as object detection. We will also generalize the cosine basis used in our method to be able to use different functions as basis that are adapted to specific scenarios or frameworks.
  
\section{Acknowledgments}
This research was supported by Secretary of Universities and Research of the Generalitat de Catalunya and the European Social Fund via a PhD grant (FI2018), and developed in the framework of project TEC2016-75976-R, financed by the Ministerio de Econom\'ia, Industria y Competitividad and the European Regional Development Fund (ERDF).

{\small
\bibliographystyle{ieee}
\bibliography{harmconv}
}

\end{document}